\pdfoutput=1
\def\year{2022}\relax
\documentclass[letterpaper]{article} 
\usepackage{aaai22}  
\usepackage{times}  
\usepackage{helvet}  
\usepackage{courier}  
\usepackage[hyphens]{url}  
\usepackage{graphicx} 
\usepackage{natbib}  
\usepackage{caption} 
\DeclareCaptionStyle{ruled}{labelfont=normalfont,labelsep=colon,strut=off} 
\usepackage{algorithm}
\usepackage{algorithmic}

\usepackage{times}
\usepackage{epsfig}
\usepackage{graphicx}
\usepackage{amsmath}
\usepackage{amssymb}
\usepackage{multirow}
\usepackage{diagbox}

\usepackage{newfloat}
\usepackage{listings}
\floatstyle{ruled}
\newfloat{listing}{tb}{lst}{}
\floatname{listing}{Listing}
\title{GTM: Gray Temporal Model for Video Recognition}

\author{
    Yanping Zhang, Yongxin Yu\thanks{Corresponding authors.}
}
\affiliations{
 
    College of Intelligence and Computing, Tianjin University, China\\
    zhangyanping210@tju.edu.cn, yyx@tju.edu.cn
}

\usepackage{bibentry}

\begin{document}

\maketitle

\begin{abstract}
	Data input modality plays an important role in video action recognition. Normally, there are three types of input: RGB, flow stream and compressed data. In this paper, we proposed a new input modality: gray stream. Specifically, taken the stacked consecutive 3 gray images as input, which is the same size of RGB, can not only skip the conversion process from video decoding data to RGB, but also improve the spatio-temporal modeling ability at zero computation and zero parameters. Meanwhile, we proposed a 1D Identity Channel-wise Spatio-temporal Convolution(1D-ICSC) which captures the temporal relationship at channel-feature level within a controllable computation budget(by parameters $G$ \& $R$).
	Finally, we confirm its effectiveness and efficiency on several action recognition benchmarks, such as Kinetics, Something-Something, HMDB-51 and UCF-101, and achieve impressive results.
\end{abstract}

\section{Introduction}

In the real world, huge amounts of video data are generated every minute. As of May 2019, more than 500 hours of video were uploaded to YouTube every minute~\cite{hale2019more}. Advances in edge computing and next generation communication technology made it possible to analyze these videos in a real time manner. So video-based task is getting more focus and becoming more important.

For video action recognition, deep learning~\cite{krizhevsky2012imagenet} has become the standard and we have witnessed great advancements.
Most of them use three types of input: RGB, optical flow and compressed data. Karpathy et al.~\cite{karpathy2014large} proposed to use a single 2D CNN model on each RGB frame independently and explored several fusing method to learn spatio-temporal features. Simonyan et al.~\cite{simonyan2014two} first proposed the two-stream networks, which included a RGB input and an optical flow~\cite{brox2004high} input respectively. Wu et al.~\cite{wu2018compressed} proposed to directly apply deep learning method in the compressed domain for action recognition. We have a question: Is there another input modality for action recognition? 

\begin{figure}[t]
	\begin{center}
		\includegraphics[width=1.0\linewidth]{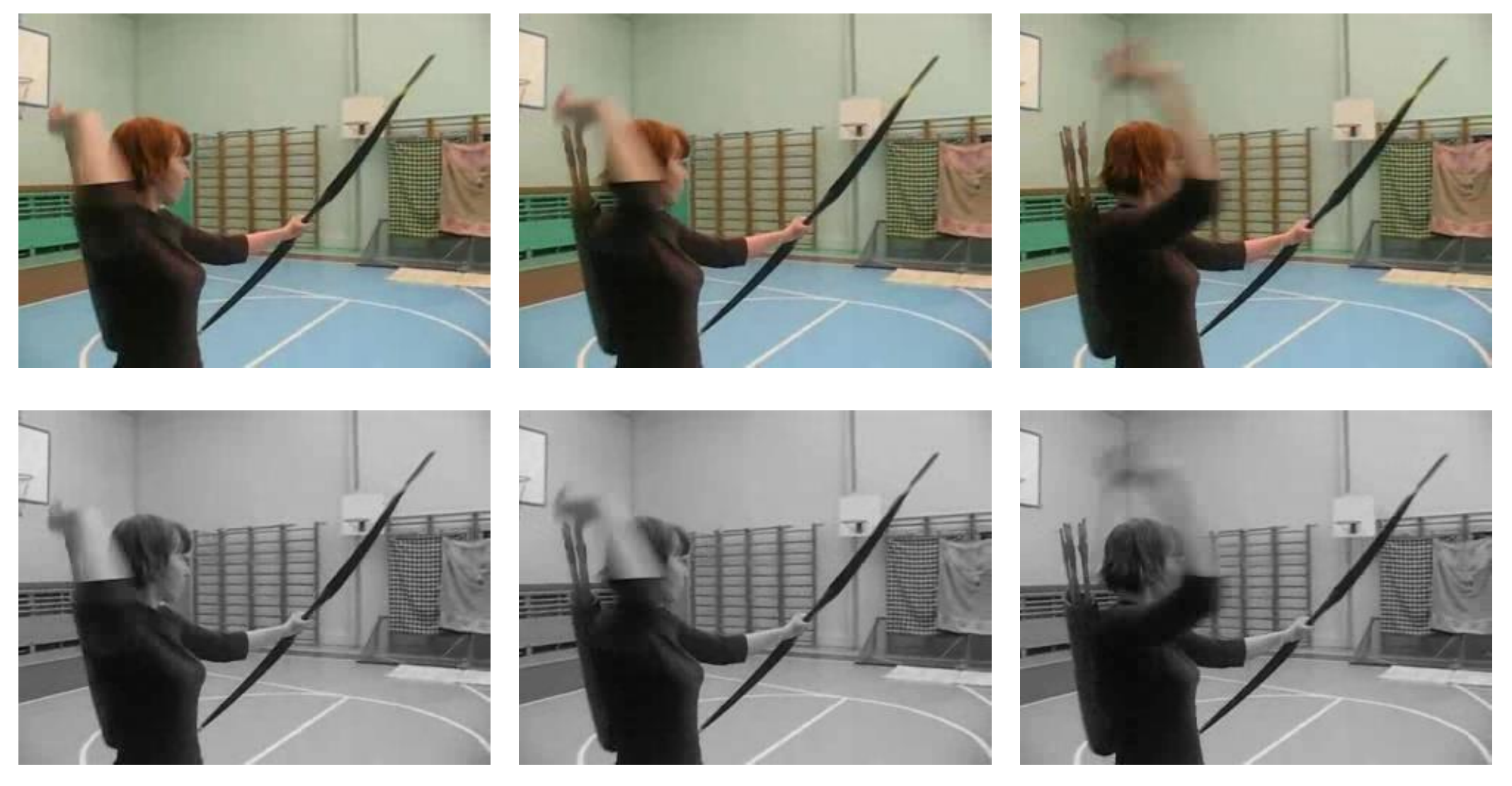}
	\end{center}
	\caption{\linespread{1.0}\selectfont Example of 3 consecutive RGB images {\itshape vs.} gray images. First row: RGB. Second row: gray.}
	\label{fig:1}
\end{figure}

\begin{figure}[t]
	\begin{center}
		\includegraphics[width=1.0\linewidth]{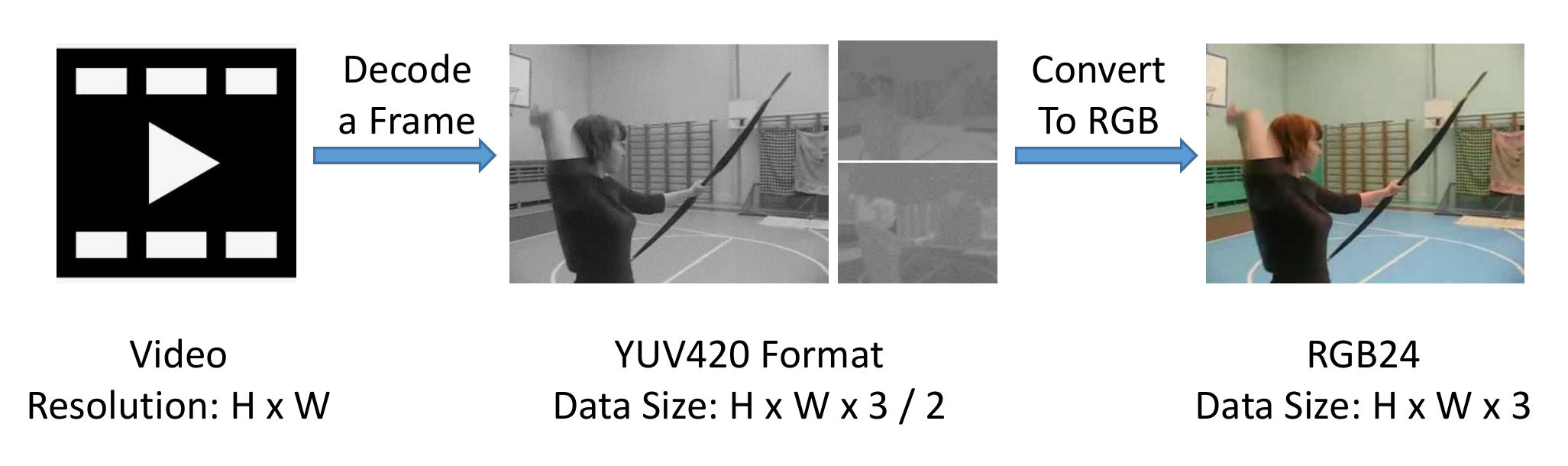}
	\end{center}
	\caption{\linespread{1.0}\selectfont A simple process of decoding a video, and then convert to RGB.}
	\label{fig:decode}
\end{figure}

The RGB format is widely used in image based deep learning methods. It is straightforward and has a large number of ready-made models, such as VGG~\cite{simonyan2014very}, Inception~\cite{szegedy2015going} and ResNet~\cite{he2016deep}. However, RGB format may not be entirely suitable for video tasks. Restricted by storage and bandwidth, video files and streams are stored or transmitted in compressed format, such as MPEG-4, H.264~\cite{wiegand2003overview}. 
After decompression, we will get YUV data directly.
Y means luminance component and UV for two chrominance components.
The YUV420 is the most widely used format which contains a subsampling process. So the data distribution of three signals(one Luma and two Chroma) is not equal.
In Figure~\ref{fig:decode}, we showed the simple process of decoding a video, and then convert to RGB format.
During the conversion from YUV420 to RGB, we observed that the data size is doubled, which requires extra computation and more storage. 

The flow stream~\cite{farneback2003two, zach2007duality} has proven to be a good representation of the short-term motion between adjacent frames. Zhao et al.~\cite{zhao2019dance} proves that the more accurate the optical flow, the more the model improves. However, the computation of optical flow is time-consuming and storage demanding, thus making it impractical for real-world deployment.

For action recognition in compressed domain, it has a long tradition~\cite{tom2013fast, ma2019effective}. The compressed data itself contains a significant number of useful clues that can be used to help classify, including Motion Vector, Residual, Quantization Parameter, Macro Block Size, MB in bits, QP Gradient.
For deep learning methods, most of them~\cite{wang2019fast, battash2020mimic} only use Motion Vector and Residual so far. And many of them are not trained and evaluated on general large-scale datasets such as kinetics~\cite{kay2017kinetics}. So the deep learning approaches in compressed domain are promising but far from being explored.

To address this issue, we investigated several video-based inputs and found that taken the stacked consecutive 3 gray images as input, which is called gray stream, can improve the modeling ability at zero computation and zero parameters. In Figure~\ref{fig:1}, we visualize the 3 consecutive RGB {\itshape vs.} gray images. The gray stream contains not only local spatial appearance information represented by individual frame but also local temporal dependency among these successive frames.

Given the new input modality, we think more about current models. 3D based CNN models involve a huge amount of computation, while 2D models lack of temporal modeling capabilities. Inspired by this observation, we propose a 1D Identity Channel-wise Spatio-temporal Convolution(1D-ICSC), which can be easily inserted into 2D CNNs with a plug-and-play manner to improve temporal modeling abilities. The 1D-ICSC learns to capture different temporal relationship for different channels at a controllable computation budget. To summarize, the main contributions of our method are three-fold:

\begin{itemize}
	\item[$\bullet$]We propose a new input modality (gray stream) for video action recognition and demonstrate its efficiency.
	
	\item[$\bullet$]A simple yet effective 1D Identity Channel-wise Spatio-temporal Convolution(1D-ICSC) is proposed, which can greatly improve temporal modeling ability for 2D based CNN.
	
	\item[$\bullet$]We evaluated the proposed method on several public benchmark datasets, 
	including Something-Something~\cite{goyal2017something}, Kinetics-400~\cite{kay2017kinetics}, UCF-101~\cite{soomro2012ucf101} and HMDB-51~\cite{kuehne2011hmdb}, and achieved impressive result.	
	
\end{itemize}

\section{Related Works}

\begin{figure*}[h]
	\begin{center}
		\includegraphics[width=0.8\linewidth]{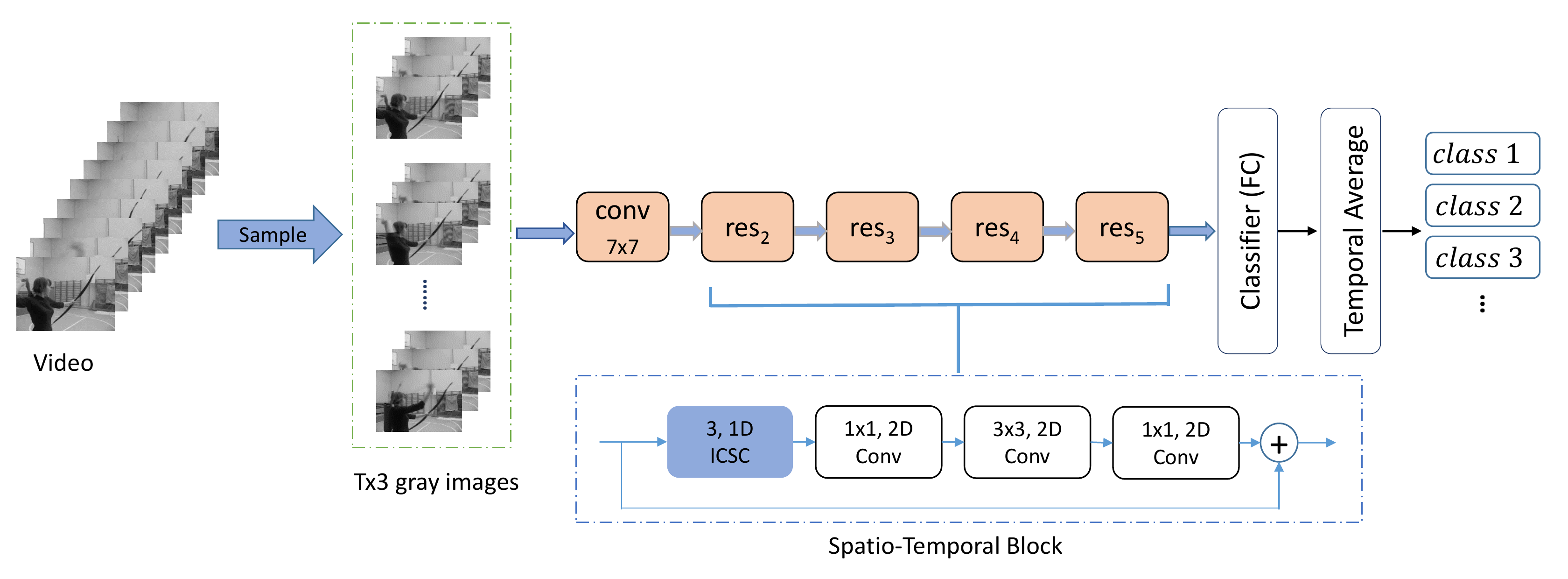}
	\end{center}
	\caption{\linespread{1.0}\selectfont Illustration of our gray temporal model with ResNet-50 backbone. One input video is divided into $T$ segments. For each segment, we random sample 3 consecutive gray images. The ResNet block is replaced by Spatio-Temporal block which has 1D ICSC inside.}
	\label{fig:mainarch}
\end{figure*}

There are several trends in video action recognition. The first one is about the network evolvement, from 2D CNNs, LSTM to 3D CNNs. Karpathy et al.~\cite{karpathy2014large} proposed to apply a 2D CNN model on large-scale Sports-1M dataset and setup the beginning of deep learning methods. 
Ng et al.~\cite{yue2015beyond} take the feature maps from CNNs then send to the LSTM network, and aggregate frame-level CNN features to model the temporal relation. In these approaches, the feature extraction of each frame is isolated and only late fusion of high-level features is performed, thus get no clear improvement. 
Tran et al.~\cite{tran2015learning} first proposed a deep 3D network, termed C3D which performed 3D convolutions on adjacent frames to jointly model the spatial and temporal features. However, with tremendous parameters to be optimized and lack of high-quality large-scale datasets, the performance of C3D remains unsatisfactory. 
The situation changed when Carreira et al.~\cite{carreira2017quo} proposed I3D, which achieved very competitive performance with the help of high-quality large-scale Kinetics~\cite{kay2017kinetics} dataset and push video action recognition to the next level. 
In the latest work Feichtenhofer introduced X3D~\cite{feichtenhofer2020x3d}, which progressively expand a tiny 2D image classification architecture along multiple network axes, such as temporal duration, spatial resolution, width, etc. X3D learned from the history of image classification models, and pushed 3D model to an extreme.

The second line mainly focuses the improvement of feature expression. Simonyan et al.~\cite{simonyan2014two} proposed the two-stream approach and setup a trend. 
Following this trend, many excellent works~\cite{feichtenhofer2016convolutional,wang2016temporal} emerged and dominated the video recognition domain from year 2014 to 2017.
Because pre-computing optical flow is computationally expensive and storage demanding, many works seeks for other substitutes.
Kantorov et al.~\cite{kantorov2014efficient} proposed the use of sparse MPEG flow instead of the dense optical flow, which improved the speed of feature extraction by two orders of magnitude with minor reduction in accuracy.

The last one focused on computational efficiency and real world deployment. ECO~\cite{zolfaghari2018eco},  TSM~\cite{lin2019tsm}, STM~\cite{jiang2019stm} and TEA~\cite{li2020tea} are the excellent ones. Lin et al. proposed a new method, termed temporal shift module(TSM). It shifts part of the channels along the temporal dimension and thus facilitate information exchange among neighboring frames. It built temporal modeling inside 2D CNNs at zero computation and zero parameters.

Among these approaches, SlowFast~\cite{feichtenhofer2019slowfast} made attempt to replace the RGB input with gray-scale input in their fast pathway. They found that the gray-scale version is nearly as good as the RGB variant, meanwhile reduces FLOPs by \%5. StNet~\cite{he2019stnet} sampled $T$ temporal segments, each of which consists of $N(N=5)$ consecutive RGB frames. These $N$ frames are stacked in the channel dimension. So the network input is a tensor of size $T \times 3N \times H \times W$ and is called super-image. Super-Image contains both spatial information and local temporal dependency. TDN~\cite{wang2021tdn} generalize the idea of RGB difference to devise an efficient temporal difference module for motion modeling. These works made remarkable research in both input modalities and network architectures.

Different from previous works, our proposed approaches focus on video-based modalities and efficient 1D spatio-temporal modeling, make it more suitable for video-based tasks and more practical for real world deployment.

\section{Approach}

In this section, we will introduce the technical details of our approach. First, we will discuss several video-based modalities, such as YCbCr. Afterward, we will present 1D-ICSC which can be embedded 2D CNN in a plug-and-play manner.
\subsection{Gray Stream}

In H.264/AVC~\cite{wiegand2003overview} as well as the previous standards(MPEG-1~\cite{international1993coding}, MPEG-2~\cite{union1994generic}), they use video color space: YCbCr\footnote {In this paper we use the terms YCbCr and YUV interchangeably, although they are not exactly the same in a strict manner.}. It separates a color representation into three components called Y, Cb, and Cr. 
Component Y is called luma, and represents brightness. The two chroma components Cb and Cr represent the extent to which the color deviates from gray toward blue and red, respectively.

\begin{figure}[t]
	\begin{center}
		\includegraphics[width=1.0\linewidth]{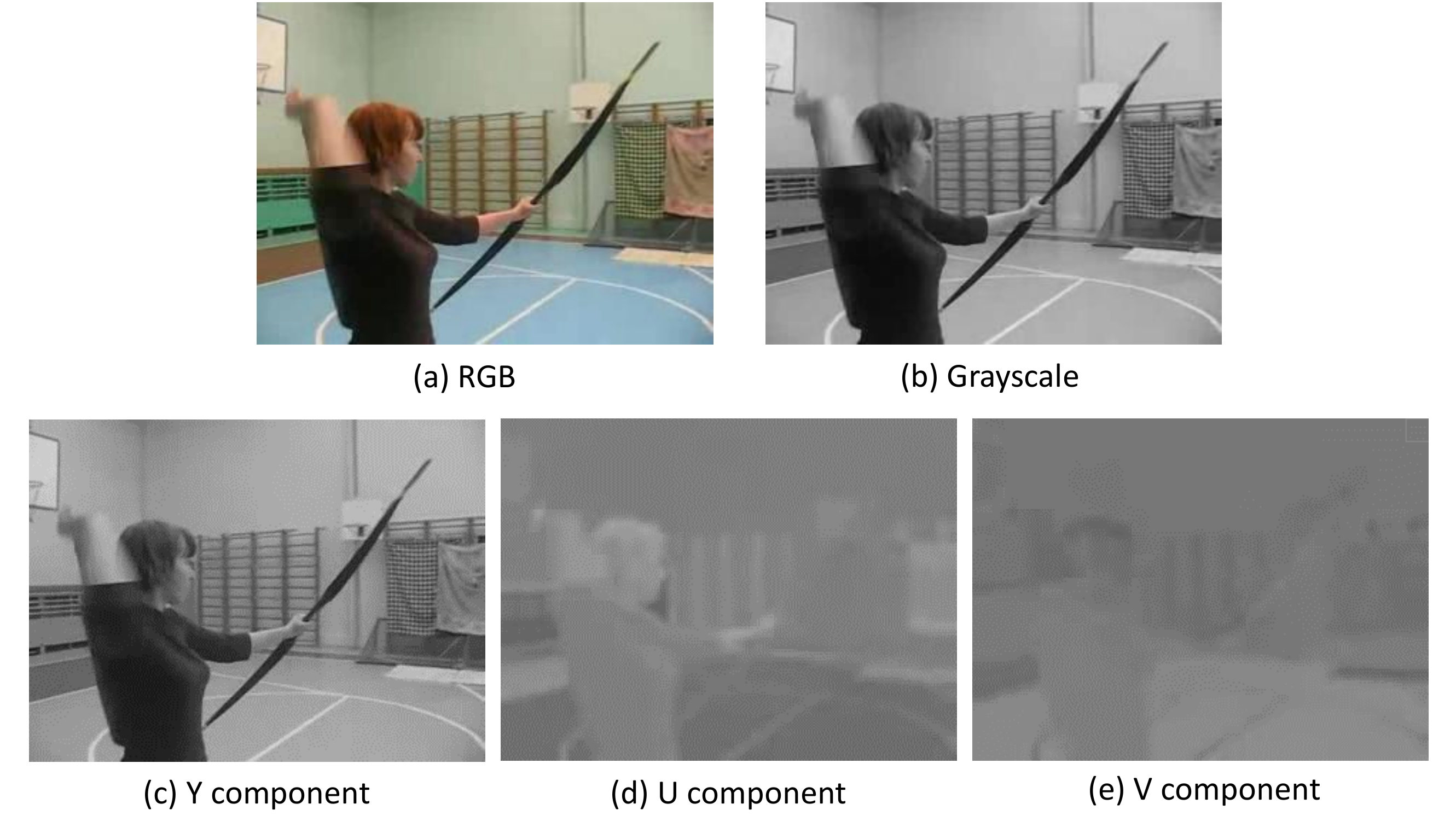}
	\end{center}
	\caption{\linespread{1.0}\selectfont Example of one RGB image and its corresponding Gray image and YUV component. The UV components are resized to the same size of Y for better visual perception.}
	\label{fig:all_modality}
\end{figure}

Because the human visual system is more sensitive to luma than chroma, subsampling is performed in which all the luma(Y) information is preserved and chroma information(CbCr) is reduced by a factor 2 in both horizontal and vertical directions. This is called 4:2:0 sampling with 8 bits of precision per sample. The whole subsampling process is lossy but does not affect the perceived quality. In Figure~\ref{fig:all_modality} we visualize the three components of YCbCr. It can be seen that a single Y component is enough for human to recognize what is going on. According to standard ITU-R Recommendation BT.601~\cite{bt2011studio}, the conversion from YCbCr to RGB is as:
\begin{align}
	& R = [Y + 1.402 \times (C_r - 128) ]_0^{255} 	\notag \\
	& G = [Y - 0.344 \times (C_b - 128) - 0.714 \times (C_r - 128) ]_0^{255} \notag \\ 
	& B = [Y + 1.772 \times (C_b - 128) ]_0^{255}  \label{YUVtoRGB}
\end{align}

{$[\quad]_{0}^{255}$} denotes clamping a value to the 8-bit range of 0 to 255. We can see that RGB frame is a transformation from YCbCr. This requires extra computation. In image domain, another widely used technology is to convert RGB image to gray-scale. According to~\cite{bt2011studio}, the conversion from RGB to gray is computed as:
\begin{align}
	& L = 0.299 \times R + 0.587 \times G + 0.114 \times B \label{RGBtoGray}
\end{align}

We visualize the gray-scale image in Figure~\ref{fig:all_modality}(b). We can see that the Y component image (Luma) and gray-scale image have visual similarity. An intuitive idea is to replace the network input with gray-scale image. So we choose gray-scale as another candidate. Until now we totally get 4 modalities: Y, U, V and Gray. All of them only have one channel, while RGB has three channels(red, green, blue). So for each modality, there are two ways to construct data. First is to use only one frame. This requires modifying the network input. Another is stacking consecutive 3 frame to form 3 channels. It is the same size of RGB and does not require any modification for network. For simplicity, we call this gray stream(for all 4 modalities). In the first ablation study, we will show its superiority over RGB and Flow.

\subsection{Spatio-Temporal Block}
\begin{figure}[ht]
	\begin{center}
		\includegraphics[width=0.7\linewidth]{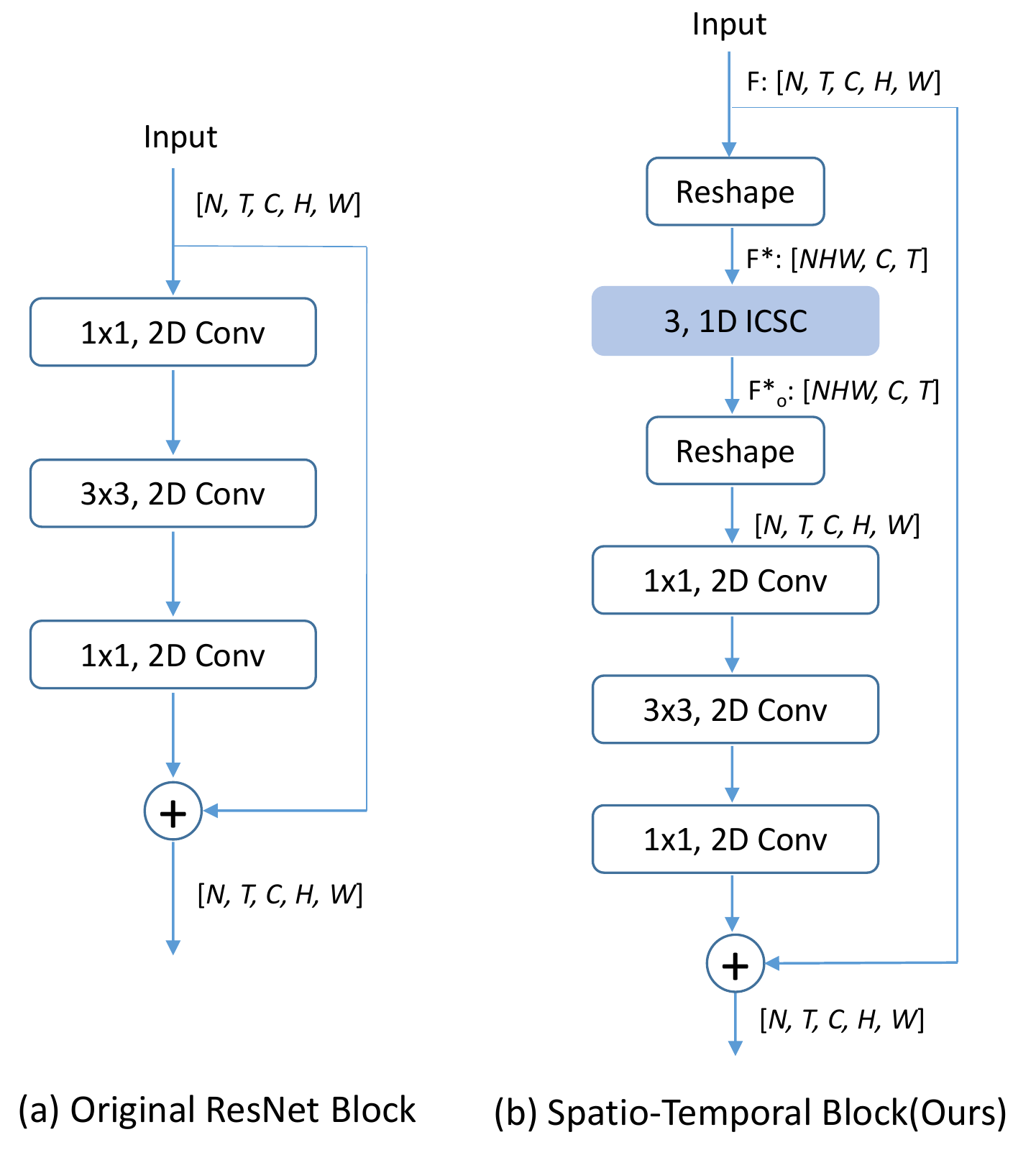}
	\end{center}
	\caption{\linespread{1.0}\selectfont (a): Original ResNet block. (b): Spatio-Temporal block. The 1D-ICSC could be easily inserted into the ResNet block to construct a Spatio-Temporal block.}
	\label{fig:channelTemporal}
\end{figure}

In order to keep the framework effective yet lightweight, we choose the TSN~\cite{wang2016temporal} with ResNet-50~\cite{he2016deep} backbone. Since a raw 2D network can not effectively capture temporal dynamics which has been evidenced by previous works~\cite{zhou2018temporal, lin2019tsm}, we designed a spatio-temporal module to tackle this problem. Figure~\ref{fig:channelTemporal} (b) shows our spatio-temporal block embedded with 1D-ICSC.

\vspace*{-4mm}
\subparagraph{1D-ICSC.} Channel-wise temporal modeling has been explored by~\cite{lin2019tsm, jiang2019stm, li2020tea} previously, which is designed to model motion information based on the channel level instead of raw pixel-level. Different from previous works, we proposed 1D-ICSC to capture the temporal relationship, and introduce two factors($G$ \& $R$) to control the computation budget. 

As illustrated in Figure~\ref{fig:channelTemporal} (b), the shape of input spatiotemporal
feature is $ \mathbf{F} \in\ \mathbb{R}^{N\times T \times C\times H\times W}$, where N is the batch size.
T and C denote temporal dimension and feature channels,
respectively. H and W correspond to spatial shape. We first reshape  $ \mathbf{F} \rightarrow \mathbf{F^*} \in\ \mathbb{R}^{NHW \times C\times T} $ and then apply the channel-wise 1D convolution as equation \eqref{1DConv}.
\begin{align}	
	\mathbf{F}^*_{o} = {}	\mathbf{K}   * \mathbf{F}^*   \label{1DConv}
\end{align}
$\mathbf{K}$ is a 1D convolutional layer with kernel size 3 and $ \mathbf{F^*_{o}} \in\ \mathbb{R}^{NHW \times C\times T} $. 
Next we reshape $ \mathbf{F^*_{o}} $ to the original input shape (i.e.[$N, T, C, H, W $]) and model local-spatial information via original ResNet block. 
Different from previous works, we specially initialized the parameters of $\mathbf{K}$ to make $\mathbf{F}^*_{o} = \mathbf{F}^* $ at initial stage. So it is called 1D Identity Channel-wise Spatio-temporal Convolution(1D-ICSC). We don't make assumptions about how the channel moves and interacts, but instead relax the kernel weights to learn it during training procedure. Experiments show that identity parameter initialization strategy brings performance improvement.

Further, we introduce two factors $G$ and $R$ to control the computation cost. $G$ is the groups number of  $\mathbf{K}$. So the weight shape of $\mathbf{K}$ is $C \times \frac{C}{G} \times 3$. $R$ is used to control how many spatio-temporal blocks are added. Theoretically, more spatio-temporal blocks will bring higher performance, but will increase parameters and FLOPs. Without loss of generality, when $\mathbf{BlockNumber~\%~R = 0}$ in each layer, we add a spatio-temporal block. Table \ref{tbl:gr_stat} shows the GFLOPs when $G$ and $R$ varies. We can see that when $G$ and $R$ increases, the FLOPs decreases.

\begin{table}[ht]
	\small
	\begin{center}
		\begin{tabular}{|c|c|c|c|c|}
			\hline
			\diagbox[width=4em]{R}{G} & 1 & 2 & 4 & 8\\
			\hline
			1 &107.26	&70.11	&51.54	&42.25 \\
			\hline
			2 &72.73	&52.85	&42.90	&37.93 \\
			\hline
			3 &57.93    &45.45	&39.20	&36.08 \\
			\hline
			4 &53.00	&42.98	&37.97	&35.47 \\		
			\hline
		\end{tabular}
	\end{center}
	\caption{The corresponding GFLOPs between $G$ \& $R$. The bigger $G$ and $R$, the smaller FLOPs. The network input is 8×3×224×224. }
	\label{tbl:gr_stat}
\end{table}

\subsection{GTM Network}

After introducing the gray stream and 1D-ICSC, we are ready to
describe how to integrate them into the existing network architecture and build the gray temporal model (GTM) network. As shown in Figure~\ref{fig:mainarch}, the 2D ResNet-50 is utilized as the backbone. First we divide one video into T segments. For each segment, we random sample 3 consecutive gray images. So the input of the network is $ {N\times T \times 3\times 224\times 224}$, which is the same size of RGB input.  From layer2 to layer5, 1D-ICSC is inserted at the beginning of each ResNet block to build Spatio-Temporal block,  which is controlled by parameter $G$ and $R$. The simple temporal pooling is applied to average action predictions for the entire video. Note that the whole framework is simple and straight forward, and does not require any modification for original blocks.

\section{Experiments}

\subsection{Dataset \& Implementations}

\subparagraph{Datasets.} We evaluate our approach on two large-scale action recognition datasets, Something-Something ~\cite{goyal2017something}, Kinetic-400~\cite{kay2017kinetics}, and other two small-scale datasets, HMDB-51~\cite{kuehne2011hmdb} and UCF-101~\cite{soomro2012ucf101}. The Something-Something V2 dataset is a large collection of humans performing actions with everyday objects. Kinetics-400 is a large-scale YouTube video dataset and we download it from CVDF~\cite{CVDF2021}, including 238,796 training videos , 19,877 validation videos and 38,671 test videos. We use Ffmpeg~\cite{ffmpeg2021} to extract the YUV data and save it in HDF5~\cite{folk2011overview}. For Kinetics, we resize the video height to 240 without changing its aspect ratio to speed up training.

\vspace*{-4mm}
\subparagraph{Models.} We choose TSM~\cite{lin2019tsm} as our baseline. To have an apple-to-apple comparison with TSM, we used the same backbone (ResNet-50) and the models are pre-trainded on ImageNet~\cite{russakovsky2015imagenet} unless stated otherwise.

\vspace*{-4mm}
\subparagraph{Training.} Most of the experimental settings are the same as TSM~\cite{lin2019tsm} and STM~\cite{jiang2019stm}.
Given an input video, we first divide it into T segments, then we randomly sample one or 3 consecutive frames from each segment.
During training, random scaling and corner cropping are utilized for data augmentation, and the
cropped region is resized to 224 × 224 for each frame. Therefore, the input size of the network is N × T × C × 224 × 224, where N is the batch size, T is the number of segments, and C is the input channel number. Horizontal flipping is applied except for Something-Something dataset.

We train our model with 2 Tesla V100(16G) GPUs. Limited by GPU memory, we set T = 8 and use a relatively small batch size 32.
For Kinetics, Something-Something v1 \& v2 datasets, the initial learning rate is 0.005. It is reduced by a factor of 10 at 30,40,45 epochs and stop at 50 epochs. The dropout rate is 0.5. 
Stochastic gradient descent (SGD) is utilized as an optimizer. Momentum and weight decay value is set to 0.9 and 1e-4. 
All the batch normalization layers~\cite{ioffe2015batch} are enabled during training.

\vspace*{-4mm}
\subparagraph{Inference.} Two evaluation protocols are considered to trade-off accuracy and speed:1) Efficient Protocol: 1-clip and center-crop where only a center crop of 224 × 224 from a single clip is used. 2) Accuracy Protocol: 10-clip and 3-crop where three crops(left, middle, right) of 224 × 224 and 10 clips are used for testing. The final prediction was the averaged score for all clips. By default we use Efficient Protocol for all tests. We only employ Accuracy Protocol for Kinetics.  
 
\subsection{Ablation Study}

\begin{table}[t]
	\small
	\begin{center}
		\begin{tabular}{@{}l@{} c@{  } c c c@{}}
			\hline
			
			Modality & Input & UCF-101 & HMDB-51 & STH-V1 \\
			\hline
			RGB 	& 8*3*H*W      & 83.1 		& 49.9 		 & 18.2 \\
			Flow 	& 8*10*H*W     & 86.6 		& {\bf 57.3} & 36.9 \\
			1-Y   	& 8*1*H*W      & 83.2 		& 47.8 	 	 & - \\
			3-Y 	& 8*3*H*W      & 87.7 		& 55.9 		 & - \\
			1-U		& 8*1*H/2*W/2  & 54.9 		& 26.7 	  	 & - \\
			3-U 	& 8*3*H/2*W/2  & 68.5 		& 37.3 	 	 & - \\
			1-V		& 8*1*H/2*W/2  & 58.2 		& 30.7 		 & - \\
			3-V 	& 8*3*H/2*W/2  & 70.1 		& 38.9 	 	 & - \\
			1-Gray 	& 8*1*H*W  	   & 83.0 		& 48.0		 & 17.3 \\
			3-Gray 	& 8*3*H*W      & {\bf 87.8} & 55.5		 & {\bf 38.9} \\
			\hline
		\end{tabular}
	\end{center}
	\caption{Comparison of {\bfseries modalities}. Input(segments * channel * height * width) denotes the input shape of one video. 1-x means one single image. 3-x means consecutive 3 images.}
	\label{tbl:modalities}
	\vspace{-1em}
\end{table}

In this section, we first conduct several ablation experiments
to testify the effectiveness of different components in our proposed methods.
The ablation experiments are performed on Something-Something v1, UCF-101 split 1 and HMDB-51 split 1. Top-1 accuracy is reported.

\subparagraph{Modalities.}
First we compare several modalities, including RGB, Grayscale, YUV and optical flow. 
We use denseflow~\cite{denseflow} to extract optical flow with Farneb{\"a}ck algorithm~\cite{farneback2003two} because of its efficiency. Here we use TSN with ResNet-50 backbone. Videos are divided into 8 segments. 
The results are shown in Table \ref{tbl:modalities}.

First for STH-V1, 3-Gray gains 20\% compared with RGB (38.9\% {\itshape vs.} 18.2\%) and it also increased by 2\% over Flow.
In UCF-101 dataset, 3-Gray modality achieved the best top-1 (87.8\%) and 3-Y also got comparable result (87.7\%). Both of them surpass Flow (86.6\%) and RGB (83.1\%).
Second, for all three datasets, 1-Gray or 1-Y achieved similar performance compared with RGB. This is consist with SlowFast~\cite{feichtenhofer2019slowfast}. It indicates for action recognition, a single Y or Gray channel contains equivalent information of RGB.
Third, U and V modalities get inferior results, we argue that UV contains less information and has lower resolution(1/2 of Y). 

In summary, compared with RGB, our gray stream can improve accuracy by a large margin without any extra parameters and FLOPs, or any optical flow pre-calculation. This indicates that video tasks are not exactly the same as image tasks.

\begin{table}[h]
	\small
	\begin{center}
		\begin{tabular}{c c c c@{}}
			\hline
			Modality				& Interval 		& UCF-101 	& STH-V1 	\\
			\hline
			\multirow{5}{*}{3-Gray} & 1 			& 87.8 		& 38.9	\\
			& 2 			& 87.1  	& 38.4	\\
			& 3 			& 87.3  	& 38.3	\\
			& 4 			& 86.3  	& 38.5	\\
			& 5 			& 86.6  	& 38.0	\\
			\hline
		\end{tabular}
	\end{center}
	\caption{Comparison of different Sampling {\bfseries Intervals}.}
	\label{tbl:ab2_interval}
\end{table}

\subparagraph{Sampling Intervals.} For our gray stream, it need 3 video frames to form one segment input. An intuitive question arises: do different sampling intervals affect the results? Here we compare different sampling intervals. 
The results are shown in Table \ref{tbl:ab2_interval}. Intervals 1 to 3 achieved similar result. From interval 4, there was a performance reduction. We argue that for 2D backbone network, the spatial modeling plays an important role, and large sampling interval will hurt the spatial modeling ability. For simplicity, we use interval 1 as default(which means 3 consecutive frames).

\begin{table}[h]
	\small
	\begin{center}
		\begin{tabular}{c c c c@{}}
			\hline
			Modalities & Backbone & UCF-101 & HMDB-51 		\\
			\hline
			
			\multirow{3}{*}{3-Y} 
			& Resnet-18 & 84.6 			& 49.0	\\
			& Resnet-34 & 86.5  		& 53.5	\\
			& Resnet-50 &  {\bf 87.7}   & {\bf 55.9}	\\
			\hline	
								
			\multirow{3}{*}{3-U} 
			& Resnet-18 & 64.9 			& 31.6	\\
			& Resnet-34 & 67.7 			& 36.1	\\
			& Resnet-50 & {\bf 68.5}  	& {\bf 37.3}	\\
			\hline
			
			\multirow{3}{*}{3-V} 	
			& Resnet-18 & 66.5 	 		& 34.9	\\
			& Resnet-34 & 69.6  		& 37.8	\\
			& Resnet-50 & {\bf 70.1}  	& {\bf 38.9}	\\
			\hline
			
			\multirow{3}{*}{3-Gray} 
			& Resnet-18 & 84.1  		& 50.7	\\
			& Resnet-34 & 86.3  		& 53.7	\\
			& Resnet-50 & {\bf 87.8}  	& {\bf 55.5}	\\
			\hline
		\end{tabular}
	\end{center}
	\caption{Comparison of different {\bfseries backbones} for 4 modalities.}
	\label{tbl:ab3_backbone}
	\vspace{-1em}
\end{table}

\subparagraph{Backbone Choice.} Because U and V image have 1/2 resolution of Y images, it is not necessary to use a relative heavy backbone Resnet-50. Here we compared 3 backbones: Resnet-18, Resnet-34 and Resnet-50 for 4 modalities. 
The results are shown in Table \ref{tbl:ab3_backbone}. For all modalities, Resnet-50 consistently achieved best results. For 3-U and 3-V, Resnet-50 provides slight performance boost(around 1\%) compared to Resnet-34. So for 3-Y and 3-Gray, we use Resnet-50 as default backbone. For 3-U and 3-V, we use Resnet-34 as default.

\begin{table}[h]
	\small
	\begin{center}
		\begin{tabular}{c c c c@{}}
			\hline
			Modality 				& Temporal 		& UCF-101 	& STH-V1 		\\
			\hline
			\multirow{6}{*}{RGB} 	
			& None 	  		& 83.1  		& 18.2	\\
			& Fixed 	  	& 83.2  		& 45.6	\\									
			& 3D-Shift    	& 83.8  		& 45.8	\\
			& 3D-Identity 	& 84.6  		& 45.9	\\
			& 1D-Shift    	& 85.0  		& 45.9	\\
			& 1D-ICSC 		& {\bf 85.3}    & {\bf 46.1}	\\
			
			\hline						
			\multirow{6}{*}{3-Gray} 
			& None		 	& 87.8  		& 38.9	\\
			& Fixed 	  	& 87.5  		& 48.8	\\									
			& 3D-Shift 		& 87.2  		& 48.7	\\
			& 3D-Identity 	& 87.9  		& 48.9	\\
			& 1D-Shift 		& 87.8  		& 48.6	\\
			& 1D-ICSC   	& {\bf 88.0}  	& {\bf 49.3}	\\
			
			\hline
			
		\end{tabular}
	\end{center}
	\caption{Comparison of different {\bfseries temporal modeling methods}. In the second column, ``None'' means no temporal modeling is used.}
	\label{tbl:ab4_1dconv}
\end{table}

\subparagraph{1D-ICSC.} Here we compare different temporal modeling methods, including Fixed, 1D Convolution and 3D Convolution. Fixed means 1/8 channels forward shift and 1/8 backward shift, which is the same as TSM~\cite{lin2019tsm}. For 1D and 3D convolution, there are two parameters initialization strategies: Identity and Shift. Identity convolutions are initialized as equation \eqref{1DConv} to make the input and output equals. Shift convolutions are initialized to perform like Fixed (1/8 channels forward and 1/8 backward). The kernel size of 3D convolution is $3 \times3 \times 3 $.

The results are shown in Table \ref{tbl:ab4_1dconv}. First we notice that Identity convolution achieved better results than Shift, both 1D and 3D convolution. Second, the 1D-ICSC achieved best result, even surpassed the 3D convolution. This indicates that proper temporal convolution is essential for temporal modeling, even though 3D convolution involves much more parameters. For STH-V1 RGB, 1D-ICSC significantly increase accuracy from 18.2\% to 46.1\%. While for 3-Gray, it also get an increase of 10\%.

\subparagraph{G \& R}
We test different $G$ \& $R$ paramters in STH-V1 dataset. The results are shown in Table \ref{tbl:ab5_rg}. Smaller $G$ \& $R$ get better results. This is consistent with Table \ref{tbl:gr_stat} as smaller $G$ \& $R$ involve more parameters and FLOPs.

\begin{table}[ht]
	\small
	\begin{center}
		\begin{tabular}{|c|c|c|c|c|}
			\hline
			\diagbox[width=4em]{R}{G} & 1 & 2 & 4 & 8\\
			\hline
			2 &50.0	& 49.6	& 49.5	& 49.2  \\
			\hline
			4 &49.3	& 49.1	& 49.1	& 49.0	 \\		
			\hline
		\end{tabular}
	\end{center}
	\caption{Top-1 accuracy under different $G$ \& $R$ paramters on STH-V1 dataset. 3-Gray modality is used.}
	\label{tbl:ab5_rg}
	\vspace{-1em}
\end{table}

\begin{table*}[t]
	\small
	\begin{center}
		\begin{tabular}{@{}  *{8}{c} @{}}
			\hline
			
			Method  & Backbone     			& Pre-train & Frames 		& Param.   &   GFLOPs   & Top-1    & Top-5\\
			\hline\hline 
			
			I3D-RGB(Carreira et al. 2017)  
			& 						&  			& 64×N/A 		&  12.7M   	& 108 × N/A   	& 71.1     & 89.3 \\ 
			I3D-Flow(Carreira et al. 2017)  
			& 3D Inception V1		& ImageNet 	& 64×N/A 		&  12.7M   	& 108 × N/A   	& 63.4     & 84.9 \\ 
			2-Stream I3D(Carreira et al. 2017) 
			& 						&  			& 128×N/A 		&  25M		& 216 × N/A		& 74.2     & 91.3 \\ 
			\hline
			
			ECO-RGB$_{En}$(Zolfaghari et al. 2018) 
			& BNIncep+3D Res18		& Scratch		& 92			& 47.5M    	& 267       & 70.0  	 & 89.4 \\
			\hline
			
			\multirow{2}{*}{NL I3D-RGB~\cite{wang2018non}} & \multirow{2}{*}{3D ResNet50} & \multirow{2}{*}{ImageNet}
			  			     		& 128     		& 35.3M    & 282		& 67.3  & -      \\
		   	&			&     		& 128×3×10    	& 35.3M    & 282×30     & 76.5  & 92.6   \\
			\hline
			
			SlowFast 8×8~\cite{feichtenhofer2019slowfast}  
			& 3D ResNet50 		 	& Scratch   	& (8+64)×3×10    & -       & 65.7×30  	& {\bf 77.0}  & 92.6 	\\
			\hline\hline
			
			\multirow{2}{*}{TSN-RGB~\cite{wang2016temporal}}			
			& BN-Inception 	 	& \multirow{2}{*}{ImageNet}    	& 25×10    		& 10.7M  	& 53×10     & 69.1	& 88.7\\
			& ResNet-50     	&     							& 8      		& 24.3M  	& 33G       & 66.8 	& -\\
			\hline
			
			R(2+1)D-RGB~\cite{tran2018closer}  & \multirow{3}{*}{ResNet-34}     & \multirow{3}{*}{Scratch} 
							    & 32×10   		& 63.8M  	& 152×10   	& 72.0	& 90.0 \\
			
			R(2+1)D-Flow~\cite{tran2018closer}       
			& 			&     	& 32×10   		& 63.8M  	& 152×10   	& 67.5	& 87.2 \\
			
			R(2+1)D 2-Stream~\cite{tran2018closer}     
			& 			&     	& 64×10   		& 127.6M  	& 304×10   	& 73.9	& 90.9 \\
			\hline

			\multirow{2}{*}{TSM~\cite{lin2019tsm}}  & \multirow{2}{*}{ResNet-50} & \multirow{2}{*}{ImageNet}      
						     	& 8   	    	& 24.3M  	& 33   		& 70.6	& - \\
			& 			&     	& 8×3×10   		& 24.3M  	& 33×30   	& 74.1	& 91.2 \\
			\hline
			
			STM-RGB~\cite{jiang2019stm}        
			& ResNet-50     		& ImageNet    	& 16×3×10   	& 24M    	& 67×30   	& 73.7	& 91.6 \\
			\hline
			
			St-Net~\cite{he2019stnet}           
			& ResNet-50     		& ImageNet    	& 25   			& 33M    	& 189   	& 69.9	& - \\
			\hline
			
			\multirow{2}{*}{TEA~\cite{li2020tea}}  & \multirow{2}{*}{Res2Net-50} & \multirow{2}{*}{ImageNet}             
			    				& 8    	 		& 24.5M      	& 35×1      & 72.5	& 90.4 \\
			&      		&     	& 8×3×10    	& 24.5M     	& 35×30     & 75.0  & 91.8 \\
			\hline
			
			TDN~\cite{wang2021tdn}          
			& ResNet-50     		& ImageNet    	& 8×3×10 		& -    		& 36×30   	& 76.6	& {\bf 92.8} \\
			\hline\hline
			
			GTM (RGB) 		&  \multirow{5}{*}{ResNet-50}     		& \multirow{5}{*}{ImageNet}    	
								& 8				& 28M   	& 43    	& 70.8  & 89.5 \\
			
			GTM (3-Y) 		
			&     		&     	& 8				& 28M   	& 43    	& 70.4 	& 89.5 \\
						
			GTM (3-Gray) 
			&     		&     	& 8				& 28M   	& 43    	& 70.4  & 89.6 \\
			
			GTM (RGB + 3-Gray) 
			&      		&     	& 16			& 49M   	& 86    	& 73.4  & 91.2 \\
			
			GTM (RGB + 3-Gray) 
			&      		&     	& 16×3×10		& 49M   	& 86×30    	& 75.2  & 92.1 \\
			
			\hline
		\end{tabular}
	\end{center}
	\caption{Comparison of our GTM network with other state-of-the-art methods on Kinetics-400 validation set.}
	\label{tbl:stoa_k400}
\end{table*}

\subsection{Comparisons with the State-of-the-arts}

In this section, we compare our proposed GTM network with the existing
state-of-the-art action recognition methods. In these experiments, we set $G$=2 and $R$=4 to get a balance between accuracy and FLOPs unless stated otherwise.

\subparagraph{Results on Kinetics-400.}
We evaluate the GTM network against the recent state-of-the-art 2D/3D
convolution-based solutions. The comprehensive statistics, including the classification results, inference protocols, parameters, and the corresponding GFLOPs, are shown in Table \ref{tbl:stoa_k400}. The first compartment contains the methods based on 3D CNNs or a mixup of 2D and 3D CNNs. The second compartment contains methods based on 2D CNNs. For fair comparison, we mainly list the architecture with ResNet-50 backbone. We can see that under Efficient Protocol, our RGB method get 70.8\%, which surpass the stNet~\cite{he2019stnet} and ECO~\cite{zolfaghari2018eco}. It is worth noting that for Kinetics dataset, flow modality usually gets inferior result than RGB. But our 3-Y and 3-Gray still achieved 70.4\%, which is comparable with RGB(70.8\%). This shows the robustness of our proposed gray stream modality. And they both surpass the flow modality methods by a large margin, such as I3D-Flow(63.4\%), R(2+1)D-Flow(67.5\%). Further a simple average of RGB and 3-Gray can bring top-1 accuracy to 73.4\%. This shows the advantage of our gray stream which is complementary to RGB.

\subparagraph{Results on STH V2.}
The Something-Something V2 dataset is more temporal-related than Kinetics. The comparison results are list in Table \ref{tbl:stoa_sthv2}. Our 3-Y achieves 61.7\% top-1 accuracy which outperforms TSM~\cite{lin2019tsm} by 2.6\%. And it also improves our RGB by 2.8\%. This indicates that in temporal-related datasets, gray stream can bring more improvement than that in scene-related datasets. The average result of RGB+3Y can increase the top-1 accuracy to 63.6\%. And when combined with 3-V, it further increases to 64.5\%. The superior performance demonstrates the effectiveness of our proposed approaches.

\begin{table}[h]
	\small
	\begin{center}
		\begin{tabular}{@{}  *{8}{c} @{}}
			\hline
			
			Method         										& GFLOPs   	&  Top-1 &  Top-5 \\
			\hline\hline 
			
			TRN Multiscale$_{8f}$~\cite{zhou2018temporal}    	& 33  		& 48.8      & 77.6 \\ 
			TRN 2-Stream$_{8f}$~\cite{zhou2018temporal}    		& -  		& 55.5      & 83.1 \\ 
			\hline
			
			TSM$_{8f}$~\cite{lin2019tsm} 			 			& 33×6 		& 59.1      & 85.6 \\ 
			\hline
			
			STM$_{8f}$~\cite{jiang2019stm} 						& 33×30  	& 62.3      & 88.8 \\ 
			\hline
			
			Dynamic$_{16f}$~\cite{wu2020dynamic} 				& 48  		& 58.2      & 85.2 \\ 
			\hline
			
			TEINet$_{8f}$~\cite{liu2020teinet} 				    & 33  		& 61.3      & - \\ 
			\hline
			
			ACTION-Net$_{8f}$(Wang et al. 2021) 				& 35  		& 62.5      & 87.3 \\
			\hline
			
			TDN$_{8f}$~\cite{wang2021tdn}  						& 36  		& 64.0      & 88.8 \\			
			\hline\hline

			GTM (RGB)		              		& 43    & 58.9        & 85.0 \\
			GTM (3-Y) 		                	& 43    & 61.7        & 87.4 \\
			GTM (3-V) 		                	& 30    & 51.6        & 80.1 \\
			GTM (RGB + 3-Y)				    	& 86    & 63.6  	  & 88.6  \\
			GTM (RGB + 3-Y + 3-V)				& 116   & {\bf 64.5}  & {\bf 89.2}  \\
			\hline
		\end{tabular}
	\end{center}
	\caption{Comparison with the state-of-the-art methods on Something-Something V2 validation set.}
	\label{tbl:stoa_sthv2}
	\vspace*{-3.5mm}
\end{table}

\subparagraph{Results on UCF-101 \& HMDB-51.}
UCF-101 and HMDB-51 are comparatively small-scale datasets with a long history, but they are still worth studying to trace the development of action recognition. We list some main results in Table \ref{tbl:stoa_ucfhmdb}. On HMDB-51 RGB, our method achieved 56.5\% compared with I3D(49.8\%). On UCF-101 RGB, our method achieved 87.4\%. When we compared 3-Y with Flow On both datasets, it surpassed Two-Stream and 3D-Fused by a large margin(around 4\%).

\begin{table}[t]
	\small	
	\begin{center}
		\begin{tabular}{@{ }l@{ } | c |c |c | c |c |c@{}}
			\hline
			& \multicolumn{3}{c|}{UCF-101}  & \multicolumn{3}{c}{HMDB-51} \\
			\hline
			Architecture  	& RGB 	& Flow 	& R+F      		& RGB 	& Flow 	& R+F \\
			\hline	
					
			LSTM*	      	& 81.0 	& - 	& -				& 36.0  & - 	& - 	\\
			\hline
			
			3D-ConvNet* 	& 51.6 	& - 	& -				& 24.3 	& -		& - 	\\
			\hline
			
			Two-Stream*    	& 83.6 	& 85.6	& 91.2			& 43.2 	& 56.3	& 58.3 	\\
			\hline
			
			3D-Fused* 	  	& 83.2 	& 85.8  & 89.3			& 49.2 	& 55.5	& 56.8 	\\
			\hline
			
			I3D*			& 84.5 	& {\bf 90.6} 	& {\bf 93.4} 	& 49.8	& {\bf 61.9} 	& {\bf 66.4}  \\
			\hline
			
			& 				& 		&  						& 		& 		& 		\\
			Architecture & RGB			& 3-Y	&  R+3Y					& RGB	& 3-Y	& R+3Y	\\
			\hline
			GTM(Ours)			& {\bf 87.4}	& 89.2	& 91.4 	& {\bf 56.5} 	& 60.2	& 62.0	\\
			\hline
		\end{tabular}
	\end{center}
	
	\caption{Comparison in UCF-101 and HMDB-51(split 1 of both). All models are pre-trained on ImageNet except 3D-ConvNet. * denotes the results are cited from I3D~\cite{carreira2017quo}. R+F means average result of RGB and Flow. R+3Y means average result of RGB and 3-Y. Here we set $G$=1 and $R$=2 for better accuracy.}
	\label{tbl:stoa_ucfhmdb}
\end{table}

\section{Conclusion}
In this paper, we proposed a new input modality gray stream for action recognition. It skips the conversion process from video decoding data to RGB, and improves the spatiotemporal modeling ability at zero computation and zero parameters. Experiments showed its superiority over RGB and Flow on various
datasets, including Kinetics-400, Something-Something, UCF-101 and HMDB-51. Further we proposed a 1D Identity Channel-wise Spatio-temporal Convolution(1D-ICSC), which is simple yet effective to improve spatio-temporal modeling ability. The further work may be to integrate gray stream and RGB into a unified framework. We hope our analysis will provide insights about video-based approaches for action recognition.

\bibliography{aaai22}

\begin{thebibliography}{49}
\providecommand{\natexlab}[1]{#1}

\bibitem[{Battash et~al.(2020)Battash, Barad, Tang, and
  Bleiweiss}]{battash2020mimic}
Battash, B.; Barad, H.; Tang, H.; and Bleiweiss, A. 2020.
\newblock Mimic the raw domain: Accelerating action recognition in the
  compressed domain.
\newblock In \emph{Proceedings of the IEEE/CVF conference on computer vision
  and pattern recognition workshops}, 684--685.

\bibitem[{Brox et~al.(2004)Brox, Bruhn, Papenberg, and Weickert}]{brox2004high}
Brox, T.; Bruhn, A.; Papenberg, N.; and Weickert, J. 2004.
\newblock High accuracy optical flow estimation based on a theory for warping.
\newblock In \emph{European conference on computer vision}, 25--36. Springer.

\bibitem[{BT et~al.(2011)}]{bt2011studio}
BT, R. I.-R.; et~al. 2011.
\newblock Studio encoding parameters of digital television for standard 4: 3
  and wide-screen 16: 9 aspect ratios.

\bibitem[{Carreira and Zisserman(2017)}]{carreira2017quo}
Carreira, J.; and Zisserman, A. 2017.
\newblock Quo vadis, action recognition? a new model and the kinetics dataset.
\newblock In \emph{proceedings of the IEEE Conference on Computer Vision and
  Pattern Recognition}, 6299--6308.

\bibitem[{CVDF(2021)}]{CVDF2021}
CVDF. 2021.
\newblock CVDF.
\newblock \url{https://github.com/cvdfoundation/kinetics-dataset/}.

\bibitem[{Farneb{\"a}ck(2003)}]{farneback2003two}
Farneb{\"a}ck, G. 2003.
\newblock Two-frame motion estimation based on polynomial expansion.
\newblock In \emph{Scandinavian conference on Image analysis}, 363--370.
  Springer.

\bibitem[{Feichtenhofer(2020)}]{feichtenhofer2020x3d}
Feichtenhofer, C. 2020.
\newblock X3d: Expanding architectures for efficient video recognition.
\newblock In \emph{Proceedings of the IEEE/CVF Conference on Computer Vision
  and Pattern Recognition}, 203--213.

\bibitem[{Feichtenhofer et~al.(2019)Feichtenhofer, Fan, Malik, and
  He}]{feichtenhofer2019slowfast}
Feichtenhofer, C.; Fan, H.; Malik, J.; and He, K. 2019.
\newblock Slowfast networks for video recognition.
\newblock In \emph{Proceedings of the IEEE/CVF International Conference on
  Computer Vision}, 6202--6211.

\bibitem[{Feichtenhofer, Pinz, and
  Zisserman(2016)}]{feichtenhofer2016convolutional}
Feichtenhofer, C.; Pinz, A.; and Zisserman, A. 2016.
\newblock Convolutional two-stream network fusion for video action recognition.
\newblock In \emph{Proceedings of the IEEE conference on computer vision and
  pattern recognition}, 1933--1941.

\bibitem[{ffmpeg(2021)}]{ffmpeg2021}
ffmpeg. 2021.
\newblock ffmpeg.
\newblock \url{https://github.com/ffmpeg/}.

\bibitem[{Folk et~al.(2011)Folk, Heber, Koziol, Pourmal, and
  Robinson}]{folk2011overview}
Folk, M.; Heber, G.; Koziol, Q.; Pourmal, E.; and Robinson, D. 2011.
\newblock An overview of the HDF5 technology suite and its applications.
\newblock In \emph{Proceedings of the EDBT/ICDT 2011 Workshop on Array
  Databases}, 36--47.

\bibitem[{for Standardization/International Electrotechnical~Commission
  et~al.(1993)}]{international1993coding}
for Standardization/International Electrotechnical~Commission, I.~O.; et~al.
  1993.
\newblock Coding of moving pictures and associated audio for digital storage
  media at up to about 1.5 Mbit/s.
\newblock \emph{ISO/IEC 11172}.

\bibitem[{Goyal et~al.(2017)Goyal, Ebrahimi~Kahou, Michalski, Materzynska,
  Westphal, Kim, Haenel, Fruend, Yianilos, Mueller-Freitag
  et~al.}]{goyal2017something}
Goyal, R.; Ebrahimi~Kahou, S.; Michalski, V.; Materzynska, J.; Westphal, S.;
  Kim, H.; Haenel, V.; Fruend, I.; Yianilos, P.; Mueller-Freitag, M.; et~al.
  2017.
\newblock The" something something" video database for learning and evaluating
  visual common sense.
\newblock In \emph{Proceedings of the IEEE International Conference on Computer
  Vision}, 5842--5850.

\bibitem[{Hale(2019)}]{hale2019more}
Hale, J. 2019.
\newblock More than 500 hours of content are now being uploaded to youtube
  every minute.
\newblock \emph{Santa Monica, CA: Tubefilter}.

\bibitem[{He et~al.(2019)He, Zhou, Gan, Li, Liu, Li, Wang, and
  Wen}]{he2019stnet}
He, D.; Zhou, Z.; Gan, C.; Li, F.; Liu, X.; Li, Y.; Wang, L.; and Wen, S. 2019.
\newblock Stnet: Local and global spatial-temporal modeling for action
  recognition.
\newblock In \emph{Proceedings of the AAAI Conference on Artificial
  Intelligence}, volume~33, 8401--8408.

\bibitem[{He et~al.(2016)He, Zhang, Ren, and Sun}]{he2016deep}
He, K.; Zhang, X.; Ren, S.; and Sun, J. 2016.
\newblock Deep residual learning for image recognition.
\newblock In \emph{Proceedings of the IEEE conference on computer vision and
  pattern recognition}, 770--778.

\bibitem[{Ioffe and Szegedy(2015)}]{ioffe2015batch}
Ioffe, S.; and Szegedy, C. 2015.
\newblock Batch normalization: Accelerating deep network training by reducing
  internal covariate shift.
\newblock In \emph{International conference on machine learning}, 448--456.
  PMLR.

\bibitem[{Jiang et~al.(2019)Jiang, Wang, Gan, Wu, and Yan}]{jiang2019stm}
Jiang, B.; Wang, M.; Gan, W.; Wu, W.; and Yan, J. 2019.
\newblock Stm: Spatiotemporal and motion encoding for action recognition.
\newblock In \emph{Proceedings of the IEEE/CVF International Conference on
  Computer Vision}, 2000--2009.

\bibitem[{Kantorov and Laptev(2014)}]{kantorov2014efficient}
Kantorov, V.; and Laptev, I. 2014.
\newblock Efficient feature extraction, encoding and classification for action
  recognition.
\newblock In \emph{Proceedings of the IEEE Conference on Computer Vision and
  Pattern Recognition}, 2593--2600.

\bibitem[{Karpathy et~al.(2014)Karpathy, Toderici, Shetty, Leung, Sukthankar,
  and Fei-Fei}]{karpathy2014large}
Karpathy, A.; Toderici, G.; Shetty, S.; Leung, T.; Sukthankar, R.; and Fei-Fei,
  L. 2014.
\newblock Large-scale video classification with convolutional neural networks.
\newblock In \emph{Proceedings of the IEEE conference on Computer Vision and
  Pattern Recognition}, 1725--1732.

\bibitem[{Kay et~al.(2017)Kay, Carreira, Simonyan, Zhang, Hillier,
  Vijayanarasimhan, Viola, Green, Back, Natsev et~al.}]{kay2017kinetics}
Kay, W.; Carreira, J.; Simonyan, K.; Zhang, B.; Hillier, C.; Vijayanarasimhan,
  S.; Viola, F.; Green, T.; Back, T.; Natsev, P.; et~al. 2017.
\newblock The kinetics human action video dataset.
\newblock \emph{arXiv preprint arXiv:1705.06950}.

\bibitem[{Krizhevsky, Sutskever, and Hinton(2012)}]{krizhevsky2012imagenet}
Krizhevsky, A.; Sutskever, I.; and Hinton, G.~E. 2012.
\newblock Imagenet classification with deep convolutional neural networks.
\newblock \emph{Advances in neural information processing systems}, 25:
  1097--1105.

\bibitem[{Kuehne et~al.(2011)Kuehne, Jhuang, Garrote, Poggio, and
  Serre}]{kuehne2011hmdb}
Kuehne, H.; Jhuang, H.; Garrote, E.; Poggio, T.; and Serre, T. 2011.
\newblock HMDB: a large video database for human motion recognition.
\newblock In \emph{2011 International conference on computer vision},
  2556--2563. IEEE.

\bibitem[{Li et~al.(2020)Li, Ji, Shi, Zhang, Kang, and Wang}]{li2020tea}
Li, Y.; Ji, B.; Shi, X.; Zhang, J.; Kang, B.; and Wang, L. 2020.
\newblock Tea: Temporal excitation and aggregation for action recognition.
\newblock In \emph{Proceedings of the IEEE/CVF Conference on Computer Vision
  and Pattern Recognition}, 909--918.

\bibitem[{Lin, Gan, and Han(2019)}]{lin2019tsm}
Lin, J.; Gan, C.; and Han, S. 2019.
\newblock Tsm: Temporal shift module for efficient video understanding.
\newblock In \emph{Proceedings of the IEEE/CVF International Conference on
  Computer Vision}, 7083--7093.

\bibitem[{Liu et~al.(2020)Liu, Luo, Wang, Wang, Tai, Wang, Li, Huang, and
  Lu}]{liu2020teinet}
Liu, Z.; Luo, D.; Wang, Y.; Wang, L.; Tai, Y.; Wang, C.; Li, J.; Huang, F.; and
  Lu, T. 2020.
\newblock Teinet: Towards an efficient architecture for video recognition.
\newblock In \emph{Proceedings of the AAAI Conference on Artificial
  Intelligence}, volume~34, 11669--11676.

\bibitem[{Ma and Song(2019)}]{ma2019effective}
Ma, M.; and Song, H. 2019.
\newblock Effective moving object detection in H. 264/AVC compressed domain for
  video surveillance.
\newblock \emph{Multimedia Tools and Applications}, 78(24): 35195--35209.

\bibitem[{Russakovsky et~al.(2015)Russakovsky, Deng, Su, Krause, Satheesh, Ma,
  Huang, Karpathy, Khosla, Bernstein et~al.}]{russakovsky2015imagenet}
Russakovsky, O.; Deng, J.; Su, H.; Krause, J.; Satheesh, S.; Ma, S.; Huang, Z.;
  Karpathy, A.; Khosla, A.; Bernstein, M.; et~al. 2015.
\newblock Imagenet large scale visual recognition challenge.
\newblock \emph{International journal of computer vision}, 115(3): 211--252.

\bibitem[{Simonyan and Zisserman(2014{\natexlab{a}})}]{simonyan2014two}
Simonyan, K.; and Zisserman, A. 2014{\natexlab{a}}.
\newblock Two-stream convolutional networks for action recognition in videos.
\newblock \emph{arXiv preprint arXiv:1406.2199}.

\bibitem[{Simonyan and Zisserman(2014{\natexlab{b}})}]{simonyan2014very}
Simonyan, K.; and Zisserman, A. 2014{\natexlab{b}}.
\newblock Very deep convolutional networks for large-scale image recognition.
\newblock \emph{ICLR}.

\bibitem[{Soomro, Zamir, and Shah(2012)}]{soomro2012ucf101}
Soomro, K.; Zamir, A.~R.; and Shah, M. 2012.
\newblock UCF101: A dataset of 101 human actions classes from videos in the
  wild.
\newblock \emph{arXiv preprint arXiv:1212.0402}.

\bibitem[{Szegedy et~al.(2015)Szegedy, Liu, Jia, Sermanet, Reed, Anguelov,
  Erhan, Vanhoucke, and Rabinovich}]{szegedy2015going}
Szegedy, C.; Liu, W.; Jia, Y.; Sermanet, P.; Reed, S.; Anguelov, D.; Erhan, D.;
  Vanhoucke, V.; and Rabinovich, A. 2015.
\newblock Going deeper with convolutions.
\newblock In \emph{Proceedings of the IEEE conference on computer vision and
  pattern recognition}, 1--9.

\bibitem[{Tom and Babu(2013)}]{tom2013fast}
Tom, M.; and Babu, R.~V. 2013.
\newblock Fast moving-object detection in H. 264/AVC compressed domain for
  video surveillance.
\newblock In \emph{2013 Fourth National Conference on Computer Vision, Pattern
  Recognition, Image Processing and Graphics (NCVPRIPG)}, 1--4. IEEE.

\bibitem[{Tran et~al.(2015)Tran, Bourdev, Fergus, Torresani, and
  Paluri}]{tran2015learning}
Tran, D.; Bourdev, L.; Fergus, R.; Torresani, L.; and Paluri, M. 2015.
\newblock Learning spatiotemporal features with 3d convolutional networks.
\newblock In \emph{Proceedings of the IEEE international conference on computer
  vision}, 4489--4497.

\bibitem[{Tran et~al.(2018)Tran, Wang, Torresani, Ray, LeCun, and
  Paluri}]{tran2018closer}
Tran, D.; Wang, H.; Torresani, L.; Ray, J.; LeCun, Y.; and Paluri, M. 2018.
\newblock A closer look at spatiotemporal convolutions for action recognition.
\newblock In \emph{Proceedings of the IEEE conference on Computer Vision and
  Pattern Recognition}, 6450--6459.

\bibitem[{Union-Telecommun(1994)}]{union1994generic}
Union-Telecommun, I.~T. 1994.
\newblock Generic coding of moving pictures and associated audio
  information-Part 2: Video.
\newblock \emph{Int. Standards Org./Int. Electrotech. Comm.(ISO/IEC) JTC 1,
  Rec. H. 262 and ISO/IEC 13 818-2 (MPEG-2 Video)}.

\bibitem[{Wang et~al.(2021)Wang, Tong, Ji, and Wu}]{wang2021tdn}
Wang, L.; Tong, Z.; Ji, B.; and Wu, G. 2021.
\newblock TDN: Temporal difference networks for efficient action recognition.
\newblock In \emph{Proceedings of the IEEE/CVF Conference on Computer Vision
  and Pattern Recognition}, 1895--1904.

\bibitem[{Wang et~al.(2016)Wang, Xiong, Wang, Qiao, Lin, Tang, and
  Van~Gool}]{wang2016temporal}
Wang, L.; Xiong, Y.; Wang, Z.; Qiao, Y.; Lin, D.; Tang, X.; and Van~Gool, L.
  2016.
\newblock Temporal segment networks: Towards good practices for deep action
  recognition.
\newblock In \emph{European conference on computer vision}, 20--36. Springer.

\bibitem[{Wang et~al.(2020)Wang, Li, Zhao, Xiong, Wang, and Lin}]{denseflow}
Wang, S.; Li, Z.; Zhao, Y.; Xiong, Y.; Wang, L.; and Lin, D. 2020.
\newblock {denseflow}.
\newblock \url{https://github.com/open-mmlab/denseflow}.

\bibitem[{Wang, Lu, and Deng(2019)}]{wang2019fast}
Wang, S.; Lu, H.; and Deng, Z. 2019.
\newblock Fast object detection in compressed video.
\newblock In \emph{Proceedings of the IEEE/CVF International Conference on
  Computer Vision}, 7104--7113.

\bibitem[{Wang et~al.(2018)Wang, Girshick, Gupta, and He}]{wang2018non}
Wang, X.; Girshick, R.; Gupta, A.; and He, K. 2018.
\newblock Non-local neural networks.
\newblock In \emph{Proceedings of the IEEE conference on computer vision and
  pattern recognition}, 7794--7803.

\bibitem[{Wiegand et~al.(2003)Wiegand, Sullivan, Bjontegaard, and
  Luthra}]{wiegand2003overview}
Wiegand, T.; Sullivan, G.~J.; Bjontegaard, G.; and Luthra, A. 2003.
\newblock Overview of the H. 264/AVC video coding standard.
\newblock \emph{IEEE Transactions on circuits and systems for video
  technology}, 13(7): 560--576.

\bibitem[{Wu et~al.(2018)Wu, Zaheer, Hu, Manmatha, Smola, and
  Kr{\"a}henb{\"u}hl}]{wu2018compressed}
Wu, C.-Y.; Zaheer, M.; Hu, H.; Manmatha, R.; Smola, A.~J.; and
  Kr{\"a}henb{\"u}hl, P. 2018.
\newblock Compressed video action recognition.
\newblock In \emph{Proceedings of the IEEE Conference on Computer Vision and
  Pattern Recognition}, 6026--6035.

\bibitem[{Wu et~al.(2020)Wu, He, Tan, Chen, Yang, and Wen}]{wu2020dynamic}
Wu, W.; He, D.; Tan, X.; Chen, S.; Yang, Y.; and Wen, S. 2020.
\newblock Dynamic inference: A new approach toward efficient video action
  recognition.
\newblock In \emph{Proceedings of the IEEE/CVF Conference on Computer Vision
  and Pattern Recognition Workshops}, 676--677.

\bibitem[{Yue-Hei~Ng et~al.(2015)Yue-Hei~Ng, Hausknecht, Vijayanarasimhan,
  Vinyals, Monga, and Toderici}]{yue2015beyond}
Yue-Hei~Ng, J.; Hausknecht, M.; Vijayanarasimhan, S.; Vinyals, O.; Monga, R.;
  and Toderici, G. 2015.
\newblock Beyond short snippets: Deep networks for video classification.
\newblock In \emph{Proceedings of the IEEE conference on computer vision and
  pattern recognition}, 4694--4702.

\bibitem[{Zach, Pock, and Bischof(2007)}]{zach2007duality}
Zach, C.; Pock, T.; and Bischof, H. 2007.
\newblock A duality based approach for realtime tv-l 1 optical flow.
\newblock In \emph{Joint pattern recognition symposium}, 214--223. Springer.

\bibitem[{Zhao and Snoek(2019)}]{zhao2019dance}
Zhao, J.; and Snoek, C.~G. 2019.
\newblock Dance with flow: Two-in-one stream action detection.
\newblock In \emph{Proceedings of the IEEE/CVF Conference on Computer Vision
  and Pattern Recognition}, 9935--9944.

\bibitem[{Zhou et~al.(2018)Zhou, Andonian, Oliva, and
  Torralba}]{zhou2018temporal}
Zhou, B.; Andonian, A.; Oliva, A.; and Torralba, A. 2018.
\newblock Temporal relational reasoning in videos.
\newblock In \emph{Proceedings of the European Conference on Computer Vision
  (ECCV)}, 803--818.

\bibitem[{Zolfaghari, Singh, and Brox(2018)}]{zolfaghari2018eco}
Zolfaghari, M.; Singh, K.; and Brox, T. 2018.
\newblock Eco: Efficient convolutional network for online video understanding.
\newblock In \emph{Proceedings of the European conference on computer vision
  (ECCV)}, 695--712.

\end{thebibliography}

\end{document}